\documentclass[conference,letterpaper]{IEEEtran}
\IEEEoverridecommandlockouts

\usepackage{cite}
\usepackage{amsmath,amssymb,amsfonts}
\usepackage{graphicx}
\usepackage{textcomp}
\usepackage{xcolor}
\usepackage{algorithm}
\usepackage{algorithmic}

\usepackage{threeparttable}
\usepackage{booktabs}
\usepackage{bm}
\usepackage{siunitx}
\usepackage{xurl}
\usepackage[hidelinks,breaklinks=true]{hyperref}
\usepackage[capitalize]{cleveref}

\makeatletter
\let\IEEEorig@makecaption\@makecaption
\long\def\@makecaption#1#2{%
  \ifx\@captype\@IEEEtablestring
    \footnotesize\bgroup\par\centering\@IEEEtabletopskipstrut
    {\normalfont\footnotesize #1:\nobreakspace\scshape #2}\par\addvspace{0.5\baselineskip}\egroup%
    \@IEEEtablecaptionsepspace
  \else
    \IEEEorig@makecaption{#1}{#2}%
  \fi}
\makeatother

\newcommand{\lam}{\bm{\lambda}}
\newcommand{\T}{^{\mathsf T}}
\newcommand{\Gdot}{\dot{G}}
\newcommand{\N}{\bm{N}}
\newcommand{\R}{\bm{R}}
\newcommand{\Phimat}{\bm{\Phi}}
\newcommand{\M}{\bm{M}}
\newcommand{\vece}{\bm{e}}

\newcommand{\alphaVec}{\bm{\alpha}}
\newcommand{\betaVec}{\bm{\beta}}

\usepackage{geometry}
\def\BibTeX{{\rm B\kern-.05em{\sc i\kern-.025em b}\kern-.08em
    T\kern-.1667em\lower.7ex\hbox{E}\kern-.125emX}}

\begin{document}

\title{Adaptive Chemotherapy Control under Tumor Heterogeneity via Reinforcement Learning}

\author{\IEEEauthorblockN{Bereket Sitotaw Kidane, Md Samiul Haque Motayed, and Shuo Wang\IEEEauthorrefmark{1}}
\thanks{This work was supported by SW's NSF CAREER Award (FAIN: 2238269). BSK, MM, and SW are with the Department of Mechanical and Aerospace Engineering, University of Texas at Arlington, Arlington, TX 76019, USA. \IEEEauthorrefmark{1}Corresponding author: SW (shuolinda.wang@uta.edu).}}

\maketitle

\begin{abstract}
Designing effective chemotherapy regimens is hindered by tumor heterogeneity and drug resistance, which complicate the deployment of patient-specific model-based optimal control across diverse populations. We develop and compare closed-loop deep reinforcement learning (DRL) dosing policies with continuous (TD3) and discrete (DQN) action spaces trained on a high-dimensional heterogeneous tumor model. The DRL policies are benchmarked against a Pontryagin's Maximum Principle (PMP)-derived open-loop benchmark. We assess generalization under parametric heterogeneity using a 100-patient virtual cohort with $\pm 10\%$ uniform perturbations in growth and drug-sensitivity parameters. Across this cohort, TD3 achieves higher average tumor reduction, while DQN yields tighter inter-patient dosing consistency, revealing a clear efficacy--consistency trade-off in this study. Our simulations assume full observation of all tumor subpopulations; translation to sparse and noisy clinical measurements will require partial-observability formulations and/or state estimation. Overall, the results show that simulation-trained DRL can learn state-dependent feedback dosing policies that complement open-loop optimal control benchmarks.
\end{abstract}


\section{Introduction}
\label{sec:introduction}
Cancer remains one of the most complex and formidable challenges in modern medicine and continues to rank among the leading causes of mortality worldwide. Its treatment paradigm is multimodal, integrating bulk tumor removal (surgery), localized control (radiation), and a growing arsenal of systemic agents such as immunotherapy and targeted therapies~\cite{hanahan2011hallmarks}. While this comprehensive approach has improved outcomes for localized tumors, the prognosis for metastatic disease remains poor, with care often shifting from curative intent to palliative management and only marginal survival extensions~\cite{dehaes1985quality}. The central barrier to cure is \emph{intra-tumor heterogeneity}: diverse cellular subpopulations driven by genetic instability evolve dynamically, fostering drug resistance and creating a moving therapeutic target~\cite{dagogo2018tumour, diaz2012tumor}. This evolutionary complexity demands a shift from static, one-size-fits-all protocols toward dynamic treatment strategies that explicitly account for tumor adaptation.

Chemotherapy remains a cornerstone of systemic therapy for metastatic cancers. The prevailing clinical paradigm, Maximum Tolerated Dose (MTD), delivers an aggressive upfront schedule to eradicate as many cancer cells as possible~\cite{fojo2009limitations}. However, insights from evolutionary biology show that this strategy can backfire: eliminating drug-sensitive cells creates an ecological vacuum that allows rare, pre-existing drug-resistant cells to flourish, a phenomenon known as competitive release~\cite{gatenby2009adaptive}. The failure of fixed protocols to account for these evolutionary dynamics motivates the design of adaptive therapies that modulate dosing to manage the tumor ecosystem and delay resistance~\cite{enriqueznavas2015eco, gluzman2020optimizing}. From a systems viewpoint, this makes dosing inherently a \emph{closed-loop} decision problem: therapy should adapt to the evolving tumor composition under toxicity constraints.

The rational design of such adaptive therapies is a complex dynamic optimization problem for which Optimal Control Theory (OCT) provides a rigorous mathematical framework~\cite{anderson2008integrative}. These tools are broadly categorized into indirect and direct methods. Indirect methods, based on Pontryagin's Maximum Principle (PMP), use variational calculus to derive necessary conditions for optimality, transforming the problem into a two-point boundary value problem~\cite{schattler2015optimal, wang2019pharmacodynamics}. While a nominal PMP solve can often be computed in seconds for a given patient model, repeated patient-specific re-optimization for closed-loop replanning as measurements arrive, or for large multi-scenario uncertainty sweeps, becomes less convenient at scale, and shooting methods exhibit sensitivity to initialization, motivating hybrid global-local solvers~\cite{kidane2025qpso}. Direct methods, such as collocation, convert the continuous optimal control problem into a large-scale nonlinear programming problem by discretizing the state and control variables~\cite{betts2010practical}. Despite their mathematical rigor, both approaches share a critical limitation in clinical settings: they are inherently model-based. Given the immense biological complexity and inter-patient variability, creating and personalizing such models with high fidelity is often infeasible, and policies optimized for a nominal model may degrade under mismatch and uncertainty~\cite{ledzewicz2013optimal}. This motivates complementary \emph{simulation-trained} approaches that learn feedback dosing policies from rollouts across heterogeneous model instances rather than relying on a single per-patient open-loop optimization.

Reinforcement learning (RL) offers such a simulation-trained, approximate dynamic programming framework, but its early applications in oncology have been hampered by a significant research gap. While demonstrating feasibility, these pioneering studies were often restricted in scope. For example, Padmanabhan et al.~\cite{padmanabhan2017reinforcement} applied tabular Q-learning to simplified pharmacological models, testing only a handful of patient scenarios. Critically, much of this foundational work relied on ad-hoc reward shaping, lacked scalability to high-dimensional models, did not benchmark performance against control-theoretic optima, and developed policies for a single nominal patient model~\cite{yu2019reinforcement, eckardt2021reinforcement}. Consequently, the resulting policies were difficult to interpret and failed to address the crucial challenge of cohort-level generalization under heterogeneity and, in practice, partial observability.

This work advances beyond these limitations by introducing a principled deep RL framework for adaptive chemotherapy. Specifically, we derive the reward signal directly from the optimal control cost functional to ensure mathematical consistency rather than relying on ad-hoc shaping. We then design and compare discrete-action (DQN) and continuous-action (TD3) agents, benchmarked against a PMP-derived solution on a nominal model to bridge the gap between model-free DRL and model-based OCT. Finally, we evaluate empirical generalization under parametric heterogeneity across a 100-patient virtual cohort, revealing a clear efficacy--consistency trade-off in our study: TD3 achieves higher mean tumor reduction, while DQN delivers greater consistency. Taken together, the paper makes three contributions: a reward formulation derived directly from the OCT cost functional, a nominal benchmark against a PMP solution, and a cohort-level robustness comparison of discrete- and continuous-action DRL policies under parametric heterogeneity. All results assume full access to the tumor subpopulation state in simulation; extending to sparse and noisy clinical measurements motivates POMDP formulations and/or state estimation. In doing so, we present a simulation-based framework that connects classical optimal control and deep RL, offering insight into adaptive chemotherapy design and a basis for future closed-loop treatment strategies under uncertainty~\cite{topol2019deep}.

\section{Problem Formulation and Optimal Control}
\label{sec:problem_and_pmp}
Chemotherapy optimization is formulated as an optimal control problem to address tumor heterogeneity, in which diverse cell subpopulations with varying drug sensitivities evolve dynamically under therapeutic pressure~\cite{greene2014density,wang2016heterogeneity}. We approximate the underlying continuum-of-traits partial differential equation (PDE) model by a finite-dimensional ordinary differential equation (ODE) system to balance biological realism and computational tractability~\cite{greene2014density}. The tumor is modeled as a collection of $n$ distinct subpopulations, represented by the state vector $\N \in \mathbb{R}^{n}$, each corresponding to a resistance trait $x_i \in [0,1]$, where $x_i=0$ denotes maximum drug sensitivity and $x_i=1$ denotes maximum resistance. The system evolves under a drug dosage $u(t)\in[0,3]$ according to
\begin{equation}
\label{eq:dynamics}
\dot{\N} = \left[ \R - \Phimat \frac{u}{1 + u} - G(\bar{N}) \M \right] \N,
\end{equation}
where $\R \in \mathbb{R}^{n \times n}$ is a diagonal matrix of replication rates, $\Phimat \in \mathbb{R}^{n \times n}$ represents cytotoxic drug effects with saturating kinetics $\frac{u}{1+u}$, and $G(\bar{N})=\log(1+\bar{N})$ models intratumoral competition, where $\bar{N}=\frac{1}{n}\sum_{i=1}^n N_i$ is the average tumor population. The matrix $\M \in \mathbb{R}^{n \times n}$ is a diagonal matrix of natural death rates~\cite{wang2019pharmacodynamics}. These parameters depend on the resistance trait $x_i$, reflecting clinically observed heterogeneous tumor responses~\cite{diaz2012tumor}.

The control objective is to choose a dose profile $u(t)$ that minimizes the cumulative cost
\begin{equation}
\label{eq:cost}
J = \alphaVec^{\mathsf T}\N(T) + \int_{0}^{T}\!\big(\betaVec^{\mathsf T}\N(t) + \gamma\,u(t)\big)\,dt,
\end{equation}
where $(\cdot)^{\mathsf T}$ denotes transpose. The vectors $\alphaVec,\betaVec\in\mathbb{R}^{n}$ weight the terminal and running tumor burdens, respectively, and $\gamma>0$ penalizes drug exposure as a toxicity proxy. Unless otherwise stated, we use $(\alpha_{\text{base}},\beta_{\text{base}},\gamma)=(500,400,4000)$, consistent with prior studies~\cite{wang2016heterogeneity}.

As a classical benchmark, we employ Pontryagin's Maximum Principle (PMP) to derive necessary conditions for an optimal control policy~\cite{schattler2015optimal}. The Hamiltonian is given by \cref{eq:hamiltonian}:
{\small
\begin{multline}
\label{eq:hamiltonian}
H(\N, u, \lam) = \betaVec\T \N + \gamma u \\
+ \lam\T \left( \R \N - \Phimat \frac{u}{1 + u}\,\N - G(\bar{N})\, \M \N \right)
\end{multline}
}
where $\lam \in \mathbb{R}^{n}$ is the vector costate, $\bar{N}=\tfrac{1}{n}\sum_{i=1}^n N_i$, and $G(\bar N)=\log(1+\bar N)$. We denote the derivative of $G$ with respect to its scalar argument by
\[
\Gdot(\bar N)\triangleq \tfrac{d}{d\bar N}G(\bar N)=\tfrac{1}{1+\bar N}.
\]

The optimal control $u^*(t)$ is obtained by solving $\frac{\partial H}{\partial u}=0$, which introduces the switching function
\[
\Psi(t)=\lam(t)\T \Phimat \N(t).
\]
The resulting control law is given by \cref{eq:optimality}:
\begin{equation}
\label{eq:optimality}
u^*(t)=\operatorname{sat}_{[0,u_{\max}]}\!\left(
\sqrt{\frac{\max(\Psi(t),0)}{\gamma}}-1
\right)
\end{equation}
where $\operatorname{sat}_{[a,b]}(x)\triangleq \min(\max(x,a),\,b)$ and $u_{\max}=3$. The costate dynamics are governed by
{\small
\begin{equation}
\label{eq:costate}
\begin{aligned}
\dot{\lam} ={}& -\betaVec
-\Big(\R-\Phimat \tfrac{u}{1+u}-G(\bar N)\M\Big)\T \lam \\
&\quad + \Gdot(\bar N)\big(\lam\T \M \N\big)\vece
\end{aligned}
\end{equation}
}
with terminal condition $\lam(T)=\alphaVec$ and $\vece=[1/n,\dots,1/n]\T$. This two-point boundary value problem is solved numerically via a shooting method~\cite{wang2019pharmacodynamics}, producing a boundary--interior control profile; see \cref{fig:pmp_dynamics}.

While PMP provides a theoretically optimal open-loop policy for a nominal patient model, its practical use is limited. The key limitation is not a single nominal solve, which can often be completed in a few seconds on a standard CPU, but the need to repeatedly re-solve a patient-specific two-point boundary value problem for replanning as new measurements arrive and for large multi-scenario uncertainty sweeps, which makes PMP less convenient at scale. Moreover, PMP relies on precise model parameters and is therefore sensitive to parametric uncertainty~\cite{diaz2012tumor}. In principle, a robust closed-loop policy could be derived from the Hamilton--Jacobi--Bellman (HJB) equation. However, for high-dimensional systems such as our $n$-subpopulation model, the HJB equation is computationally intractable because of the curse of dimensionality~\cite{schattler2015optimal, bertsekas2019reinforcement}. This dual challenge---PMP's reliance on brittle models and the intractability of robust theoretical solutions based on HJB---motivates a computational framework that can handle this dimensionality. DRL addresses this gap by framing the problem as a Markov decision process (MDP) and using deep neural networks to approximate the Bellman optimality equation, the discrete-time counterpart of the HJB~\cite{sutton2018rl}.

\section{Deep Reinforcement Learning Framework}
\label{sec:drl_methodology}
Building on the need to overcome the HJB equation's intractability, we frame the chemotherapy problem as a Markov Decision Process (MDP)~\cite{sutton2018rl,bertsekas2019reinforcement}. The MDP defines the state $s = \N$ (tumor subpopulations), action $a = u$ (drug dose), and reward as the negative incremental cost from \cref{eq:cost}, $r = -(\betaVec\T \N(t) + \gamma u(t))\,dt$, with state transitions governed by \cref{eq:dynamics}. Episodes start from a prescribed (or cohort-sampled) initial condition $\N(0)$, and the terminal penalty in \cref{eq:cost} is included via $r_T=-\alphaVec\T \N(T)$. We assume full observation of all $n=21$ tumor subpopulations, i.e., $s=\N$; this is an idealization. In practice, only sparse and noisy biomarkers may be available, motivating POMDP formulations and/or state estimation as future work~\cite{kaelbling1998planning}. The optimal action-value function $Q^*(s,a)$ satisfies the Bellman optimality equation (\cref{eq:bellman})~\cite{sutton2018rl,bertsekas2019reinforcement}, which DRL learns to approximate from simulated trajectories under parametric variations. We selected Deep Q-Network (DQN) for its stability and alignment with discretized dosing protocols~\cite{fahrenbruch2018dose,mnih2015human}, and Twin-Delayed DDPG (TD3) for its robust, fine-grained continuous control in nonlinear dynamics~\cite{fujimoto2018addressing}. Unlike prior single-patient DRL studies~\cite{yu2019reinforcement,padmanabhan2017reinforcement,eckardt2021reinforcement}, our dual approach benchmarks these agents against a PMP-derived benchmark and quantifies cohort-level generalization under parametric heterogeneity across 100 virtual patients.
\begin{align}
\label{eq:bellman}
Q^*(s,a) = \mathbb{E}\!\left[r + \gamma_d \max_{a'} Q^*(s', a') \mid s, a\right],
\end{align}
where $\gamma_d\in(0,1]$ is the RL discount factor, $s'$ is the next state, and $\mathbb{E}[\cdot]$ denotes expectation over transition randomness during training~\cite{sutton2018rl,bertsekas2019reinforcement}.

\subsection{Deep Q-Network (DQN) for Discrete Control}
DQN approximates the optimal action-value function $Q^*(s,a)$ using a neural network $Q_\theta(s,a)$~\cite{mnih2015human}. To stabilize learning, it employs two key components outlined in Alg.~\ref{alg:dqn}: an experience replay buffer $\mathcal{D}$ and a separate, delayed target network $Q_{\theta'}$~\cite{mnih2015human}. During training, transitions are collected using an $\epsilon$-greedy policy and stored in $\mathcal{D}$; replay sampling breaks temporal correlations and improves data efficiency~\cite{mnih2015human}. For each sample, a stable one-step temporal-difference (TD) target is computed using the frozen target network, as shown in \cref{eq:dqn_target}~\cite{mnih2015human}:
\begin{equation}
y = r + \gamma_d \max_{a'} Q_{\theta'}(s', a'),
\label{eq:dqn_target}
\end{equation}
and the network parameters are updated by minimizing the squared TD-error (MSBE) loss~\cite{mnih2015human,sutton2018rl}:
\begin{align}
\label{eq:dqn_loss}
L(\theta) = \mathbb{E}_{(s,a,r,s') \sim \mathcal{D}} \left[ \left( y - Q_{\theta}(s, a) \right)^2 \right].
\end{align}
Here $(s,a,r,s')\sim\mathcal{D}$ denotes a transition sampled from the replay buffer, and the expectation in \cref{eq:dqn_loss} is approximated by an empirical average over a mini-batch.

\begin{algorithm}[!t]
\caption{DQN Training Algorithm}
\label{alg:dqn}
{\footnotesize
\begin{algorithmic}[1]
\STATE Init $Q_\theta$, target $Q_{\theta'}\leftarrow Q_\theta$, replay $\mathcal{D}$~\cite{mnih2015human}.
\FOR{steps}
\STATE $a\sim\epsilon$-greedy$(Q_\theta)$; store $(s,a,r,s')$ in $\mathcal{D}$; sample batch~\cite{mnih2015human}.
\STATE $y=r+\gamma_d\max_{a'}Q_{\theta'}(s',a')$; update $\theta$ by MSBE~\cite{mnih2015human,sutton2018rl}; if $t\bmod C=0$ set $\theta'\!\leftarrow\!\theta$.
\STATE $s\leftarrow s'$.
\ENDFOR
\end{algorithmic}}
\end{algorithm}

\subsection{Twin-Delayed DDPG (TD3) for Continuous Control}
For continuous dose control, we use TD3, an actor-critic method that learns a deterministic actor policy $\pi_\phi(s)$ and two critic networks ($Q_{\theta_1}, Q_{\theta_2}$)~\cite{fujimoto2018addressing}. TD3 mitigates Q-value overestimation by integrating three techniques: (i) clipped double Q-learning via the pessimistic $\min$ of two critics, (ii) target policy smoothing by adding bounded noise to the target action, and (iii) delayed policy updates~\cite{fujimoto2018addressing}. Accordingly, the TD3 target is given by \cref{eq:td3_target}~\cite{fujimoto2018addressing}:
\begin{equation}
\label{eq:td3_target}
y = r + \gamma_d \min_{i=1,2} Q_{\theta'_i}\!\left(s', \pi_{\phi'}(s') + \epsilon\right),
\end{equation}
where $\epsilon=\mathrm{sat}_{[-c,c]}(\xi)$, $\xi\sim\mathcal{N}(0,\tilde{\sigma}^2)$, and
$\mathrm{sat}_{[-c,c]}(x)\triangleq \min\{\max\{x,-c\},\,c\}$.
Here $\epsilon$ implements target-policy smoothing, and the $\min$ operator reduces overestimation bias by using the more conservative of the two target critics~\cite{fujimoto2018addressing}.

\begin{algorithm}[!t]
\caption{TD3 Training Algorithm}
\label{alg:td3}
{\footnotesize
\begin{algorithmic}[1]
\STATE Init actor $\pi_\phi$, critics $Q_{\theta_1},Q_{\theta_2}$, targets $(\phi',\theta_1',\theta_2')$, replay $\mathcal{D}$~\cite{fujimoto2018addressing}.
\FOR{steps}
\STATE Act $a=\pi_\phi(s)+\eta$; store $(s,a,r,s')$; sample batch~\cite{fujimoto2018addressing}.
\STATE $\tilde a'=\pi_{\phi'}(s')+\mathrm{sat}_{[-c,c]}(\xi)$, $\xi\sim\mathcal{N}(0,\tilde{\sigma}^2)$; $y=r+\gamma_d\min_i Q_{\theta_i'}(s',\tilde a')$; update critics~\cite{fujimoto2018addressing}.
\STATE If $t\bmod d=0$: update actor (DPG) and Polyak-update targets~\cite{fujimoto2018addressing}; set $s\leftarrow s'$.
\ENDFOR
\end{algorithmic}}
\end{algorithm}

\section{Numerical Results}
\label{sec:exp_setup_results}
We evaluate the deep reinforcement learning (DRL) state-feedback policies for the system in \cref{eq:dynamics} under the cost structure of \cref{eq:cost} in two stages. First, we benchmark DQN and TD3 against the PMP solution on a nominal patient model. Second, we assess robustness by deploying one trained policy from each agent across a 100-patient virtual cohort with parametric uncertainty.
\begin{figure}[!t]
\centering
\includegraphics[width=\columnwidth]{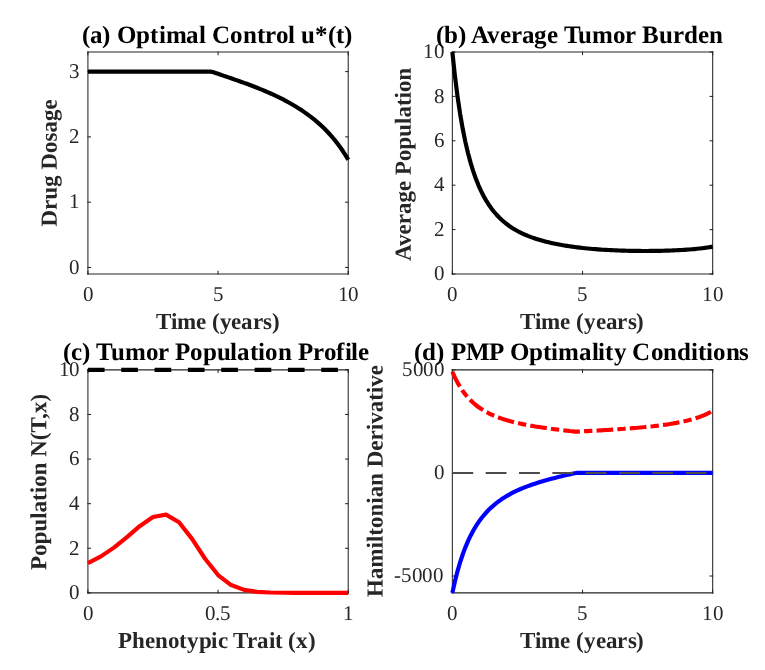}
\caption{Optimal policy derived from PMP for the nominal patient. (a) The bang-singular control profile (black solid line). (b) Resulting tumor burden over time (black solid line). (c) Initial (black dashed line) and final (red solid line) tumor distributions. (d) Hamiltonian derivatives satisfying optimality, where $\partial H/\partial u$ (blue solid line) is zero during singular arcs and the second derivative $\partial^2 H/\partial u^2$ (red dash-dot line) is positive.}
\label{fig:pmp_dynamics}
\end{figure}

\begin{figure}[!t]
\centering
\includegraphics[width=\columnwidth]{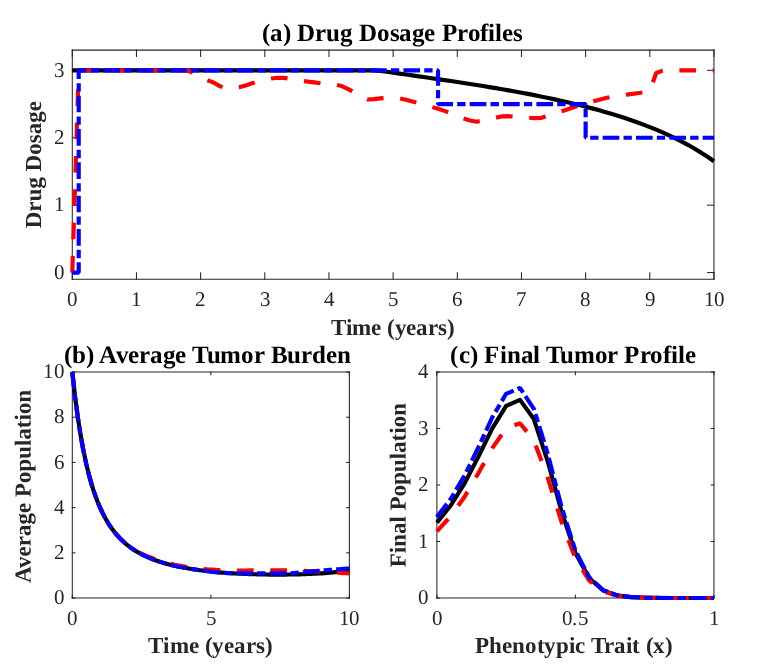}
\caption{Performance comparison on the nominal patient model. (a) Dosage profiles and (b) tumor burden reduction for the PMP benchmark (black solid line), continuous TD3 agent (red dashed line), and discrete DQN agent (blue dash-dot line).}
\label{fig:nominal_trajectory}
\end{figure}

For all simulations, we use the $n=21$-dimensional ODE model in \cref{eq:dynamics}, with resistance traits $x_i=(i-1)/20$, $i=1,\dots,21$. Following~\cite{pahnehkolaei2023optimal}, we set $\R=\mathrm{diag}[2/(1+3x_i^4)]$, $\Phimat=\mathrm{diag}[-\sin(x_i-1)+1.5]$, and $\M=0.5\,I_{21}$. The control is bounded by $u_{\max}=3$ over a 10-year horizon with $dt=0.1$, and all state inputs are normalized by $N_{\text{norm}}=10.0$. For DQN, the action space $u\in[0,3]$ is discretized into 7 levels based on the ablation study in \cref{tab:dqn_ablation}. For robustness testing, the nominal growth ($\R$) and drug-sensitivity ($\Phimat$) parameters were perturbed by multipliers drawn from $\mathcal{U}[0.9,1.1]$. The agents were implemented in MATLAB R2024b using the Reinforcement Learning Toolbox\texttrademark{} with hyperparameters listed in \cref{tab:drl_params}; updates follow \cref{eq:dqn_target,eq:td3_target}. Both agents use two-layer feedforward networks (400 and 300 neurons) with ReLU activations and batch normalization. Training converged after about 3,000 episodes, requiring roughly 2.5 hours for DQN and 1.5 hours for TD3 on a 2.3~GHz 8-Core Intel Core i9 CPU with 32~GB RAM.
\begin{figure}[!t]
\centering
\includegraphics[width=\columnwidth]{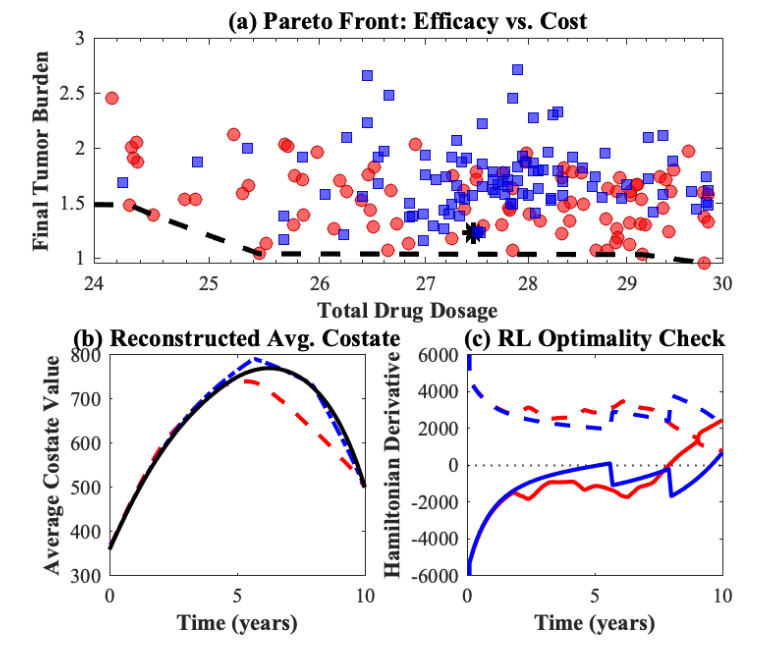}
\caption{Policy analysis across the 100-patient cohort. (a) Pareto plot of final outcomes (PMP: black star, TD3: red circles, DQN: blue squares, Pareto front: black dashed line). (b) Reconstructed average costates (PMP: black solid, TD3: red dashed, DQN: blue dashed line). (c) Hamiltonian derivatives for the optimality check (TD3: red lines, DQN: blue lines for first derivative [solid] and second derivative [dashed]).}
\label{fig:rl_analysis}
\end{figure}

\begin{figure}[!t]
\centering
\includegraphics[width=\columnwidth]{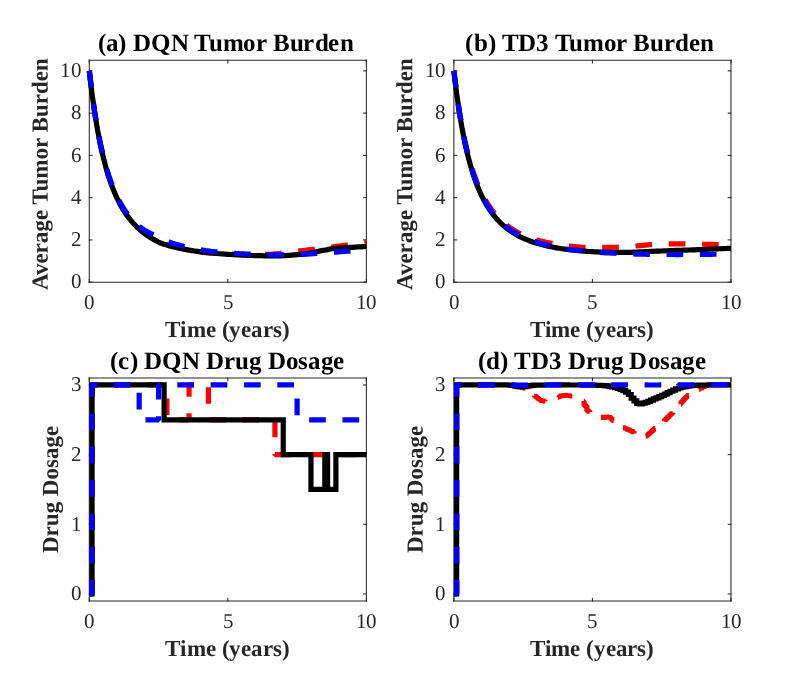}
\caption{Robustness trajectories across the patient cohort showing median (black solid), best-case (red dashed), and worst-case (blue dashed) outcomes. (a)-(b) Tumor burden evolution for DQN and TD3 policies. (c)-(d) Corresponding adaptive drug dosage profiles, with DQN's discrete actions shown as step functions.}
\label{fig:robust_trajectory}
\end{figure}
On the nominal patient model, the PMP-derived policy defined by \cref{eq:optimality}, together with the state dynamics in \cref{eq:dynamics}, provides the optimal open-loop benchmark (\cref{fig:pmp_dynamics}), achieving an 87.7\% tumor reduction. As summarized in \cref{tab:nominal_performance} and visualized in \cref{fig:nominal_trajectory}, both DRL agents learn closed-loop policies that successfully approximate this benchmark. Notably, the TD3 agent achieves a slightly higher tumor reduction than the computed PMP benchmark in the discretized simulation setting (89.1\% vs.\ 87.7\%). This result suggests that the learned state-feedback policy can exploit online state dependence within the simulation, rather than indicating that TD3 surpasses the theoretical optimum of the underlying continuous-time optimal control problem. To provide a deeper, control-theoretic analysis of these learned policies, we performed an optimality check based on Pontryagin's Maximum Principle. This diagnostic process began by reconstructing the costate trajectory, $\lam(t)$, associated with each DRL agent's simulated state and control trajectories, $\N(t)$ and $u(t)$, as shown in \cref{fig:rl_analysis}(b). This was achieved by integrating the costate differential equation in \cref{eq:costate} backward in time from the known terminal condition $\lam(T)=\alphaVec$, using the simulated state trajectories generated by \cref{eq:dynamics}. With the complete set of trajectories ($\N(t)$, $u(t)$, $\lam(t)$) in hand, we then calculated the analytical derivatives of the Hamiltonian in \cref{eq:hamiltonian} with respect to the control at each time step. As presented in \cref{fig:rl_analysis}(c), for a policy to be optimal, the first derivative, $\partial H / \partial u$, should be near zero (unless the control is saturated), and the second derivative, $\partial^2 H / \partial u^2$, must remain positive. This diagnostic check indicates how closely the DRL policies align with these necessary conditions for optimality, but it should not be interpreted as a proof of optimality.

With nominal performance validated, we deployed the single trained policies to the 100-patient cohort. The performance trade-offs are visualized in the Pareto plot (\cref{fig:rl_analysis}(a)), which reveals the core efficacy-versus-consistency trade-off. The TD3 agent demonstrates superior efficacy; its outcomes (red circles) define the Pareto front and achieve better average performance, but with higher variance. Conversely, the DQN agent offers superior consistency; its outcomes (blue squares) are tightly clustered, indicating a more predictable and reliable response to patient variability.

Numerically, as detailed in \cref{tab:robustness_performance}, TD3 achieves a statistically significant increase in mean tumor reduction relative to DQN (84.07\% vs.\ 82.78\%, Wilcoxon rank-sum test: $p=0.0078$; standardized mean difference, Cohen's $d=0.4$), along with a lower final tumor burden (1.59 vs.\ 1.72 units). However, DQN's discrete policy yields lower outcome variability, evidenced by a much tighter 95\% confidence interval on its total drug dosage ($\pm$0.23 for DQN vs. $\pm$0.40 for TD3). The resulting adaptive policies for both agents, shown in \cref{fig:robust_trajectory}, mitigate resistant rebound in these simulations, highlighting the practical trade-off between higher average efficacy and greater consistency under modeled heterogeneity.


\begin{table}[!t]
\centering
\caption{DQN Action Space Ablation Study}
\label{tab:dqn_ablation}
\sisetup{round-mode=places, round-precision=3}
\setlength{\tabcolsep}{3.5pt} 
\begin{tabular}{@{}lS[round-precision=2, table-format=2.2]S[table-format=1.3]S[round-precision=2, table-format=2.2]S[table-format=1.3e+1]@{}}
\toprule
\textbf{Agent} & {\textbf{Tumor}} & {\textbf{Final}} & {\textbf{Total}} & {\textbf{Total}} \\
\textbf{Config.}& {\textbf{Red. (\%)}}& {\textbf{Burden}}& {\textbf{Drug}} & {\textbf{Cost}} \\
\midrule
DQN (4 Act.)   & 87.04 & 1.31 & 27.00 & 2.894e5 \\
DQN (7 Act.)   & 86.93 & 1.31 & 26.70 & 2.888e5 \\
DQN (13 Act.)  & 85.37 & 1.40 & 25.74 & 2.930e5 \\
\bottomrule
\end{tabular}
\end{table}

\begin{table}[!t]
\centering
\caption{DRL Agent Hyperparameters}
\label{tab:drl_params}
\begin{tabular}{lcc}
\toprule
\textbf{Hyperparameter} & \textbf{DQN Value} & \textbf{TD3 Value} \\
\midrule
Optimizer & Adam & Adam \\
Actor Learning Rate & -- & 1.5e-4 \\
Critic Learning Rate & 5e-5 & 5e-5 \\
Discount Factor ($\gamma_d$) & 0.995 & 0.995 \\
Buffer Size & 1e6 & 1e6 \\
Mini-Batch Size & 256 & 256 \\
Use Double DQN & True & -- \\
Target Smooth Factor ($\tau$) & 1e-3 & 1e-3 \\
Target Policy Noise & -- & 0.3 \\
Policy Update Delay & -- & 2 \\
Exploration Method & $\epsilon$-Greedy & Gaussian Noise \\
Initial $\epsilon$ / Noise $\sigma$ & 1.0 & 0.5 \\
Decay Rate & 1e-4 & 1e-6 \\
\bottomrule
\end{tabular}
\end{table}

\begin{table}[!t]
\centering
\caption{Nominal Performance Benchmark}
\label{tab:nominal_performance}
\sisetup{round-mode=places,round-precision=2}
\begin{tabular}{lccc}
\toprule
\textbf{Controller} & \textbf{Final Tumor} & \textbf{Reduction (\%)} & \textbf{Total Drug} \\
\midrule
PMP (Optimal) & 1.23 & 87.72 & 27.50 \\
DQN & 1.31 & 86.93 & 26.70 \\
TD3 & 1.09 & 89.11 & 26.90 \\
\bottomrule
\end{tabular}
\end{table}

\begin{table}[!t]
\centering
\caption{Robustness Performance Across 100-Patient Cohort}
\label{tab:robustness_performance}
\sisetup{round-mode=places,round-precision=2}
\begin{tabular}{@{}lcc@{}}
\toprule
\textbf{Metric} & \textbf{DQN} & \textbf{TD3} \\
\midrule
\multicolumn{3}{l}{\textit{Mean $\pm$ 95\% CI}} \\
Tumor Reduction (\%) & $82.78 \pm 0.61$ & $84.07 \pm 0.67$ \\
Final Tumor Burden & $1.72 \pm 0.06$ & $1.59 \pm 0.07$ \\
Total Drug Dosage & $27.65 \pm 0.23$ & $27.13 \pm 0.40$ \\
\addlinespace
\multicolumn{3}{l}{\textit{Quartile Analysis (Median [25th, 75th])}} \\
Tumor Reduction (\%) & 83.2 [81.2, 84.7] & 84.0 [82.2, 86.7] \\
Total Drug Dosage & 27.65 [27.42, 27.88] & 27.13 [26.73, 27.53] \\
\bottomrule
\end{tabular}
\end{table}

\section{Discussion}
\label{sec:discussion}

The central finding of this study, summarized in \cref{tab:robustness_performance}, is a clear \emph{efficacy--consistency trade-off} under the modeled biological uncertainty. The continuous-action TD3 policy achieves higher mean efficacy (greater average tumor reduction, $p=0.0078$), making it an attractive choice when maximizing therapeutic impact is the primary objective. In contrast, the discrete-action DQN policy exhibits lower outcome variance across the cohort, improving consistency and predictability. This dichotomy suggests a practical interpretation: TD3’s higher mean efficacy may be attractive when average tumor reduction is prioritized, whereas DQN’s lower variance may be preferable when dosing predictability and consistency are emphasized; these examples are illustrative rather than clinical recommendations~\cite{gallagher2024mathematical}.

From a control-theoretic perspective, the reconstructed costate trajectories in \cref{fig:rl_analysis}(b) help explain this trade-off. DQN’s quantized action space is more aligned with bang--singular structure and yields lower-magnitude costate excursions (e.g., $|\lambda_{\text{DQN}}| < 1.2 \|\lambda_{\text{PMP}}\|$ on average), consistent with more conservative dose adjustments. TD3, unconstrained by discretization, learns smoother state-feedback policies that can adjust dosing earlier in response to evolving tumor composition. This additional flexibility may improve mean performance under parameter variations while respecting the dose bounds $u(t)\in[0,u_{\max}]$.

This approximate dynamic programming (ADP) framework~\cite{lewis2009reinforcement}
provides a complementary feedback alternative to classical open-loop optimization. A single offline training phase ($\sim$2 hours on a standard CPU) yields a state-feedback policy $u(\mathbf{N}(t))$ that is evaluated online at negligible cost. By contrast, PMP can often compute a nominal open-loop schedule in seconds for a given patient model \emph{when a suitable costate initialization/continuation is available}, but shooting-based solves are sensitive to initialization and may fail to converge otherwise; moreover, PMP must be re-solved for closed-loop replanning (as new measurements arrive) or for large multi-scenario uncertainty sweeps. In our simulations, the learned feedback policy adapts dosing to the evolving tumor composition and may reduce resistance-driven rebound associated with competitive release relative to a fixed open-loop schedule. All results assume full observation of the $n=21$ tumor subpopulations ($\mathbf{s}(t)=\mathbf{N}(t)$), which is an idealization; in practice, clinical measurements are typically sparse and noisy, motivating natural extensions using POMDP formulations and/or state estimation. For example, recent work formulates sequential clinical decision-making as a POMDP and incorporates explicit belief updates (e.g., HMM-based filtering), coupled with an optimal-stopping objective solved via deep RL~\cite{wang2024earlydiag}. Similarly, future work could investigate explicit safety-filtering mechanisms for treatment constraints.

\section{Conclusion}
\label{sec:conclusion}
This work introduced a DRL framework for \emph{adaptive} chemotherapy scheduling and benchmarked closed-loop policies (DQN and TD3) against a PMP-derived open-loop benchmark on a high-dimensional heterogeneous tumor model.
We further evaluated empirical generalization under parametric heterogeneity across a 100-patient virtual cohort.
A key finding is a consistent efficacy--consistency trade-off in our simulations: TD3 achieves higher mean tumor reduction, while DQN yields tighter dosing consistency. Future work will address partial observability through POMDP-based formulations and state-estimation-based feedback design, for example by combining reduced clinical measurements with observer-based state reconstruction.

\bibliographystyle{ieeetr}
\bibliography{mylib}

\end{document}